\documentclass{article}

\usepackage{arxiv}

\usepackage[utf8]{inputenc} 
\usepackage[T1]{fontenc}    
\usepackage{hyperref}       
\usepackage{url}            
\usepackage{booktabs}       
\usepackage{amsfonts}       
\usepackage{nicefrac}       
\usepackage{microtype}      
\usepackage{xcolor}         
\usepackage{graphicx}
\usepackage{multirow}
\usepackage{amsmath}
\usepackage{multicol}

\title{\texttt{Ingredients}: Blending Custom Photos with \\ Video Diffusion Transformers}

\author{%
Zhengcong Fei, \  Debang Li,  \ Di Qiu
\\
\textbf{Changqian Yu\thanks{Corresponding author}, \   Mingyuan Fan} \\
Kunlun Inc.\\
Beijing, China\\
{\tt\small \{feizhengcong\}@gmail.com}
}

\begin{document}

\maketitle

\begin{abstract}
  This paper presents a powerful framework to customize video creations by incorporating multiple specific identity (ID) photos, with video diffusion Transformers, referred to as \texttt{Ingredients}. Generally, our method consists of three primary modules: (\textbf{i}) a facial extractor that captures versatile and precise facial features for each human ID from both global and local perspectives; (\textbf{ii}) a multi-scale projector that maps face embeddings into the contextual space of image query in video diffusion transformers; (\textbf{iii}) an ID router that dynamically combines and allocates multiple ID embedding to the corresponding space-time regions. Leveraging a meticulously curated text-video dataset and a multi-stage training protocol, \texttt{Ingredients} demonstrates superior performance in turning custom photos into dynamic and personalized video content. Qualitative evaluations highlight the advantages of proposed method, positioning it as a significant advancement toward more effective generative video control tools in Transformer-based architecture, compared to existing methods. The data, code, and model weights are publicly available as GitHub. 
\end{abstract}

\section{Introduction}

Recent advancements in large-scale pre-trained video diffusion models have reformed the field of text-to-video synthesis \cite{hong2022cogvideo,blattmann2023stable,singer2022makeavideo,girdhar2023emuvideo}, especially the scaling property of diffusion transformers architecture \cite{peebles2023dit,opensora_plan, opensora, allegro, yang2024cogvideox,movie_gen,kong2024hunyuanvideo}, enabling a wide range of downstream applications \cite{magictime, controlnet, InstaDrag, evagaussians, cycle3d, ViewCrafter}. These controllable scenes are particularly impactful in the area of customization \cite{dreamvideo, customvideo, motionbooth, Still-moving, magic-me, fang2024motioncharacter}, with a notable emphasis on multi-human personalization, due to its vast potential in domains like AI-generated portraits, video animation, and multi-scene storyboard \cite{xing2024survey}. The primary aim to produce video contents that preserve consistent facial identity (ID) across multiple reference images, while simultaneously incorporating additional user-defined prompts.

\begin{figure*}[t]
  \centering
   \includegraphics[width=0.99\linewidth]{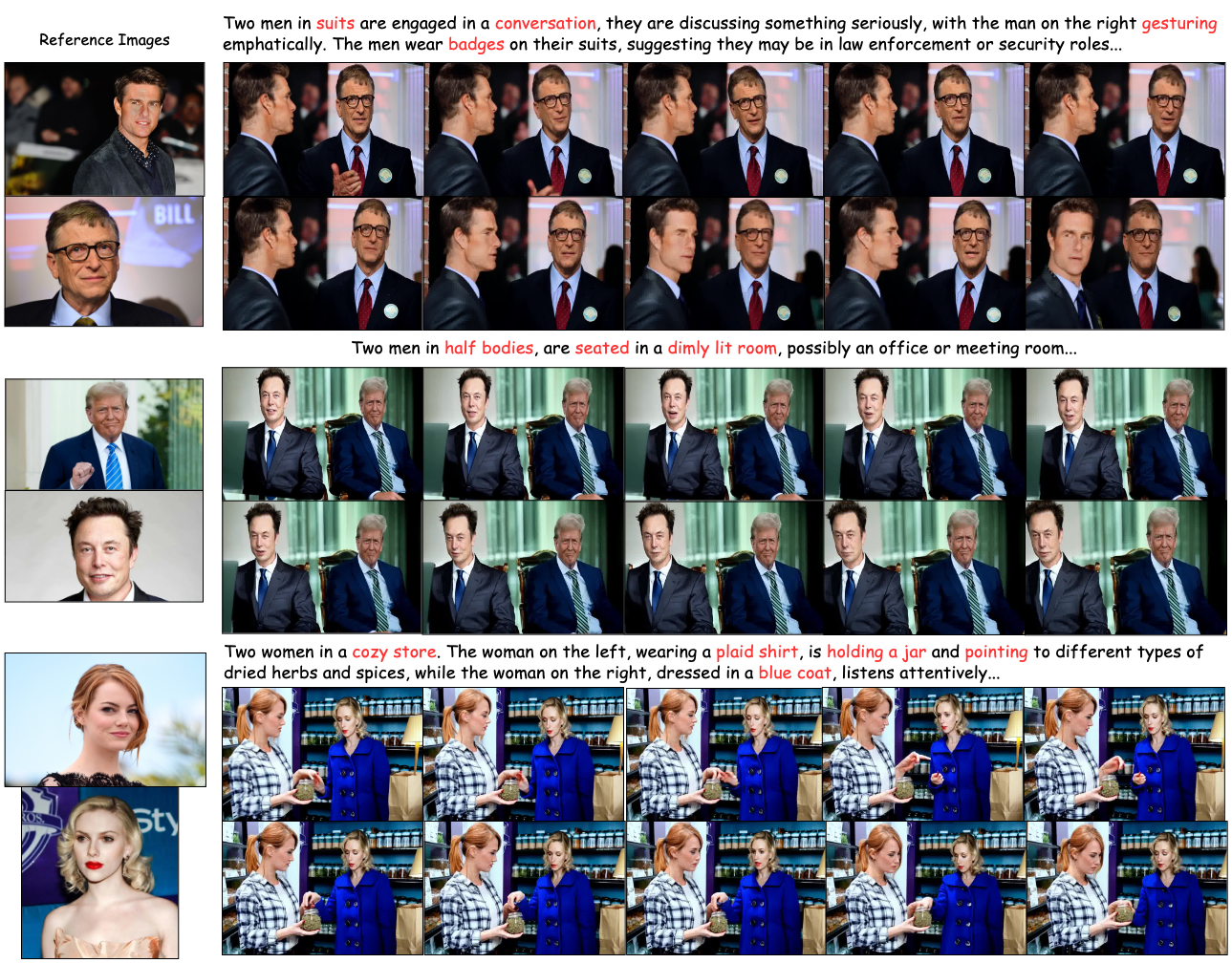}
   \caption{\textbf{Examples of multi-ID customized video results from our proposed \texttt{Ingredients}.} Given a reference with multiple human image set, our method can generate realistic and personalized videos while preserving specific human identity consistent. 
   }
   \label{fig:case1} 
\end{figure*}

However, existing personalization approaches encounter several challenges. 
For training-based methods, overhead associated with case-by-case fine-tuning limits their scalability and broader applicability \cite{Dreambooth,textual_inversion}. For training-free methods, they are predominantly based on U-Net \cite{ren2024customize,wu2024customcrafter,dreamvideo,Videobooth} and cannot be adapted to the emerging DiT-based video generative model \cite{opensora, allegro, easyanimate,yang2024cogvideox,kong2024hunyuanvideo}. Among the few exceptions, ConsisID \cite{yuan2024identity} can integrate with video diffusion transformers, but it is restricted to single-human customization. This challenge is likely due to the fundamental differences between DiT and U-Net \cite{rethinking_DiT, ViT, U-ViT, U-DiTs, yu2024promptfix,fei2024dimba,fei2024scaling}, such as the greater difficulty in training convergence and weakness in perceiving facial details. 
Notably, commercial video generation models like Keling and Vido have made significant strides in maintaining multi-ID consistency. Our work aims to bridge this gap within the open-source community, striving for similar advances in multi-ID customization for video diffusion.

In this work, we introduce a solution for multi-ID video customization with video diffusion transformers, termed as \texttt{Ingredients}, aims to achieve high-fidelity identity preservation, enhanced content flexibility, and natural video generation. Generally, \texttt{Ingredients} comprises three key components: (\textbf{i}) a facial extractor that extracts versatile editable facial features for each ID from global and local perspectives; (\textbf{ii}) a multi-scale projector that projects embeddings to the context space of image query;(\textbf{iii}) an ID router that dynamically allocates and integrates the ID embeddings across their respective regions.
The multi-stage training process sequentially optimizes the ID embedding and ID router components. As a result, our method enables the customization of multiple IDs without prompt constraints, offering great adaptability and precision in video synthesis.
Extensive experiments validate the superiority of our proposed method over existing techniques, highlighting its effectiveness as a generative control tool. More notably, our approach offers the flexibility to customize every aspect of the video creation process, making it well-suited for diverse applications such as personal storytelling, promotional videos, and creative projects. It ensures that each generated video is uniquely tailored, aligned with the user’s vision, and retains a personalized touch. 

The contributions are summarized as below: 
\begin{itemize}
    \item We propose \texttt{Ingredients}, a training-free multi-ID customization framework based on video diffusion transformers, that ensures the preservation of multiple IDs in generated videos while supporting precise textual control signals; 
    \item We present a routing mechanism that combines and distributes multiple ID embeddings in the context space of image query dynamically. Supervised with classification loss,  it avoids blending of various identities and preserving individuality;
    \item We design a multi-stage training process that optimizes face embedding extraction and multi-ID routing in sequence, resulting in enhanced facial fidelity and improved controllability of generated visual content; 
    \item Extensive experiments demonstrate the proposed framework can produce high-quality, editable, and consistent multi-human customized videos qualitatively and quantitatively, showing superiority performance with baselines. 
    To support continued exploration, we have made our data, models, and training configurations publicly available.
\end{itemize}

\section{Related Works}

\subsection{Text-to-Video Generation}

Conditional video generation has undergone substantial advancements, transitioning from initial methodologies based on GANs and VAEs to more sophisticated frameworks such as diffusion models. Early GAN-based methods \cite{vondrick2016generating, saito2017temporal, tulyakov2018mocogan, clark2019adversarial, yu2022generating} encountered challenges related to temporal coherence, leading to inconsistencies between consecutive frames. To address these issues, video diffusion models incorporating U-Net architectures—originally developed for text-to-image generation—have been adapted, improving frame continuity.
In recent developments, diffusion Transformers \cite{peebles2023dit, fei2024scaling,fei2024flux,fei2023masked} and their variants \cite{U-ViT, fei2024scalable} have replaced the traditional convolutional U-Net backbone \cite{ronneberger2015u} with pure Transformer architectures.
They usually \cite{peebles2023dit, lu2023vdt, ma2024latte, gao2024lumina} use spatio-temporal attention and full 3d attention, which enhanced the ability to capture the intricate dynamics of video and ensure consistent frame transitions \cite{he2022latent, blattmann2023align, chen2023seine, girdhar2023emuvideo}.
This paradigm shift has led to notable improvements in scalability and simplified parameter expansion.  
Meantime, auto-regressive models \cite{yan2021videogpt, hong2022cogvideo, villegas2022phenaki, kondratyuk2023videopoet, xie2024show, liu2024mardini,fei2024diffusion,fei2019fast,fei2021partially} have also employed with discrete tokens to effectively model temporal dependencies, particularly excelling in the generation of long-form videos \cite{yin2023nuwa, wang2023genlvideo, zhao2024moviedreamer, henschel2024streamingt2v, tan2024videoinfinity, zhou2024storydiffusion}.  
Our study explores the controllable application of video diffusion Transformers, with a focus on multi-ID customization.

\subsection{ID-Preserving Video Generation}

Large-scale, pre-trained text-to-video models \cite{esser2021taming, esser2024scaling} have leveraged advanced text embeddings \cite{radford2021learning, raffel2019exploring} and sophisticated attention mechanisms, thereby advancing research in fine-grained control over the video generation process \cite{li2025controlnet, mou2024t2i, bao2023latentwarp, qiu2024moviecharacter, peng2024controlnext,fei2024video,fei2023jepa}. 
In the context of ID preservation, tuning-based methods requires fine-tuning pre-trained model for each new individual during inference \cite{Dreambooth,hu2021lora,textual_inversion,fei2023gradient,fei2022deecap}. Although subsequent developments have led to the introduction of both image and video models based on U-Net and DiT architectures \cite{Hyperdreambooth, multi-concept, dreamvideo, customvideo, motionbooth, Still-moving, ID-Animator, fang2024motioncharacter}, the cost to continual fine-tune for each new identity restricts their scalability and practical applicability. 
To address the issue of high resource consumption, several tuning-free methods have recently emerged in the field of image generation \cite{Ip-adapter, instantid, UniPortrait, pulid, photomaker}. However, in the domain of video, only ConsisID \cite{yuan2024identity} currently support ID-preserving text-to-video generation with DiT. 
Note that Keling, Vido, and MovieGen \cite{movie_gen} are closed-source, whereas ConsisID is open-source but only support single human customization. 
Here we continued select the advanced video diffusion transformers architecture \cite{yang2024cogvideox} and optimize it for multi-ID scenarios, enables enabling the generation of high-quality, editable, and consistent identity-preserving videos.

\section{Methodology}

\begin{figure*}[t]
  \centering
   \includegraphics[width=0.9\linewidth]{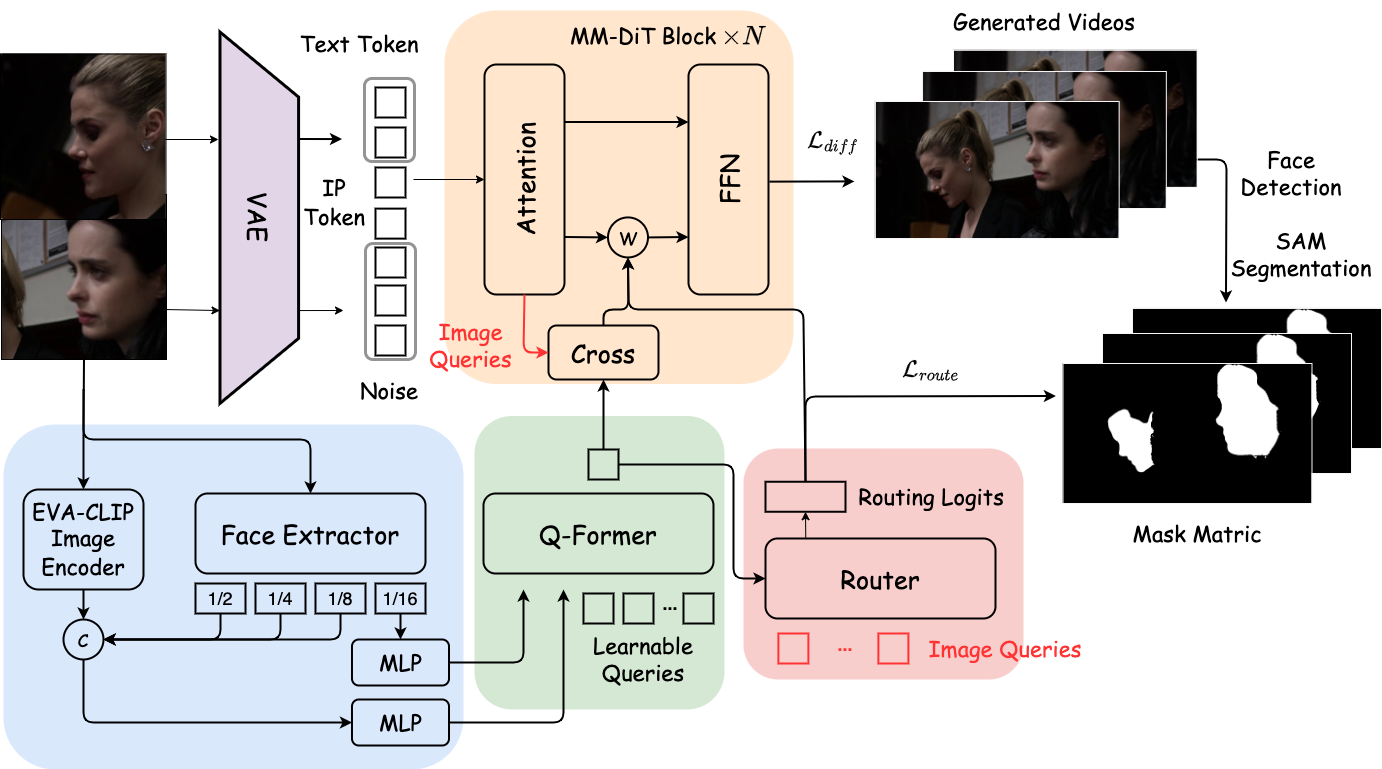}
   \caption{\textbf{Overview of \texttt{Ingredients} framework.} The proposed method consists of three key modules: a facial extractor, a q-former-based projector, and an ID router. 
   The facial extractor collects versatile editable facial features with a decoupling strategy for each ID. 
   The q-former projector map multi-scale facial embedding into different layers of video diffusion transformers. 
   The ID router combines and distributes ID embeddings to their respective locations adaptively without the intervention for prompts and layouts. 
   The entire training process of the framework is curated into two stages, i.e., the facial embedding alignment stage and the router fine-tuning stage.
   }
   \label{fig:framework} 
\end{figure*}

\subsection{Preliminaries}

\paragraph{Diffusion Model.}
Text-to-video generative models usually adopt the diffusion paradigm, which incrementally transforms a noise $\epsilon$ into a video $x_0$. Early approaches performed the denoising process directly within the pixel space \cite{DDIM, DDPM, DPM}. However, due to the substantial computational demands, more recent techniques predominantly leverage latent spaces \cite{LDM, kondratyuk2023videopoet, magictime, yang2024cogvideox,kong2024hunyuanvideo} with 3D VAE. The optimization objective can be defined as:
\begin{equation}
\label{eq: original_loss_function}
{L}_{diff}=\mathbb{E}_{x_0, t, y, \epsilon}\left[\left\|\epsilon-\epsilon_\theta\left(\mathcal{E}\left(x_0\right), t, \tau_\theta(y)\right)\right\|_2^2\right],
\end{equation}
where $y$ denotes textual prompt, $\epsilon$ is randomly sampled from a Gaussian distribution, \emph{e.g.}, $\epsilon \sim \mathcal{N}(0, 1)$, and $\tau_\theta(\cdot)$ is the text encoder, e.g., T5 \cite{raffel2019exploring}. By substituting $x_0$ with its latent representation $\mathcal{E}\left(x_0\right)$, form the diffusion process, which is also used in our method.

\paragraph{Diffusion Transformers.}
Video generation models based on DiT demonstrates considerable promise in modeling the dynamics of the physical world \cite{SORA, yang2024cogvideox, allegro}. Although this architecture represents a recent innovation by scaling with 1D sequence, research on controllable remains under-explored. Existing techniques \cite{tora, pulid, Camera_DiT, movie_gen} primarily mirror the design principles of U-Net-based models \cite{controlnet, t2iadapter, textual_inversion}. Notably, no prior study has investigated the mechanisms underlying the efficacy of DiT in scenarios of multi-ID customization. Conversely, commercial models have attained highly impressive results. This work aims to bridge the disparity between open-source solutions and their proprietary counterparts.

\subsection{Key Modules}

The proposed \texttt{Ingredients}, which is capable of generating identity-preserving videos given multiple reference images, mainly consists of three principal modules: a facial extractor, and q-former multi-scale projector, and an ID router, as illustrated in Figure \ref{fig:framework}.

\paragraph{Facial Extractor.}

The facial extractor is meticulously designed to encode high-fidelity, editable facial identity information of given images, enabling the video diffusion transformers to produce identity-consistent and controllable video outputs. In alignment with \cite{pulid,yuan2024identity}, the module employs both global and local perspectives to derive facial features:
\begin{itemize}
    \item The \textbf{global facial embedding} process begins with face detection, where individual face regions with boxes are extracted from the image set and assembled into a single large composite image. To mitigate distortions caused by resizing or the loss of facial details from cropping, a padding operation using white pixels ensures that the final composite adheres to the required dimensions. This synthesized image, containing multiple identities, is subsequently input into a VAE to extract shallow feature representations.
    \item For \textbf{local facial embeddings}, a face recognition backbone is utilized to extract features that robustly represent intrinsic identity attributes. Simultaneously, a CLIP \cite{radford2021learning} image encoder is employed to capture semantically rich features. It is important to note that each identity within the image set retains independent features, i.e., for two reference images, we maintain two sets of local facial embeddings.
\end{itemize}

\paragraph{Multi-scale Projector.}

To effectively map each facial embedding into the context space of image queries from the video diffusion transformers, we address two scenarios:
\begin{itemize}
    \item For \textbf{global facial embedding}, which consolidates all identities into a single composite image following VAE processing, is directly concatenated with the latent noise input;
    \item For each \textbf{local facial embedding} set $F^n$, $n=1, \ldots, N$, and $N$ is the number human ID, a multi-step fusion is adopted. Initially, multi-scale features extracted from the facial recognition backbone are concatenated with the corresponding CLIP-derived features. Subsequently, a Q-former structure \cite{pulid} employing cross-attention is utilized to enable interaction between these features and the visual tokens from the video diffusion transformer's attention blocks. This process can be mathematically expressed as:
    \begin{equation}
        \tilde{F}^{n} = \text{Cross-Attention}(Q_i^{img}, K_i^{face}, V_i^{face}),
    \end{equation}
    where $Q_i^{img}=H_i^{img} W^q_i$, $K_i^f = F^n W_i^k$, $V_i^f = F^n W_i^v$, $H_i^{img}$ is the queried visual token from $i$-th attention block, and $W_q, W_k, W_v$ are trainable parameters.
\end{itemize}

\begin{figure*}[t]
  \centering
   \includegraphics[width=0.99\linewidth]{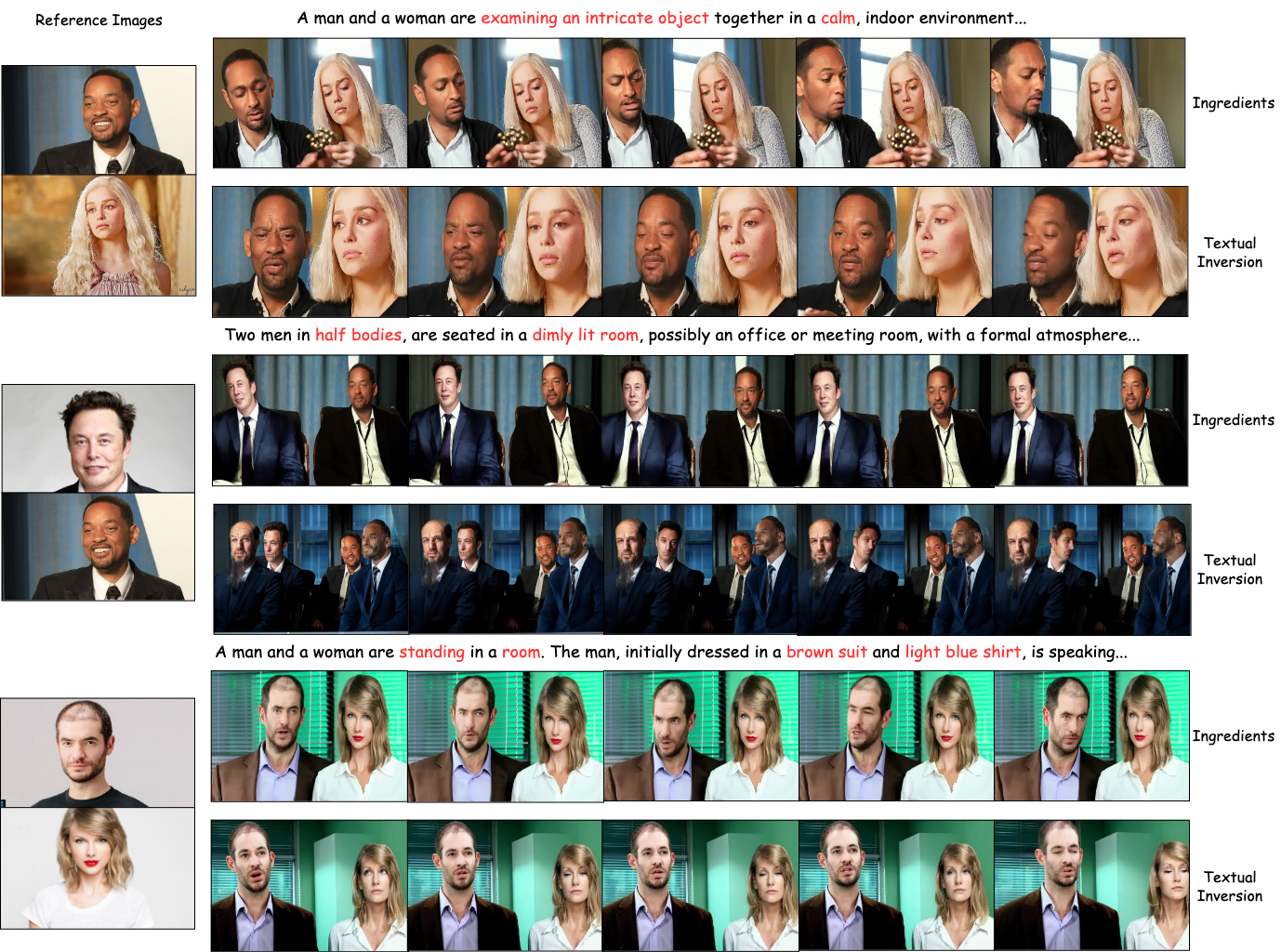}
   \caption{\textbf{Qualitative comparison of different personalization methods on multi-ID video customization.} It can been seen that compared with training-based customization, i.e., textual inversion, our method can clearly routing and attention the respect regions, benefits to ID consistency as well as strong prompt following.  
   }
   \label{fig:compare_tt} 
\end{figure*}

\begin{figure*}[t]
  \centering
   \includegraphics[width=0.99\linewidth]{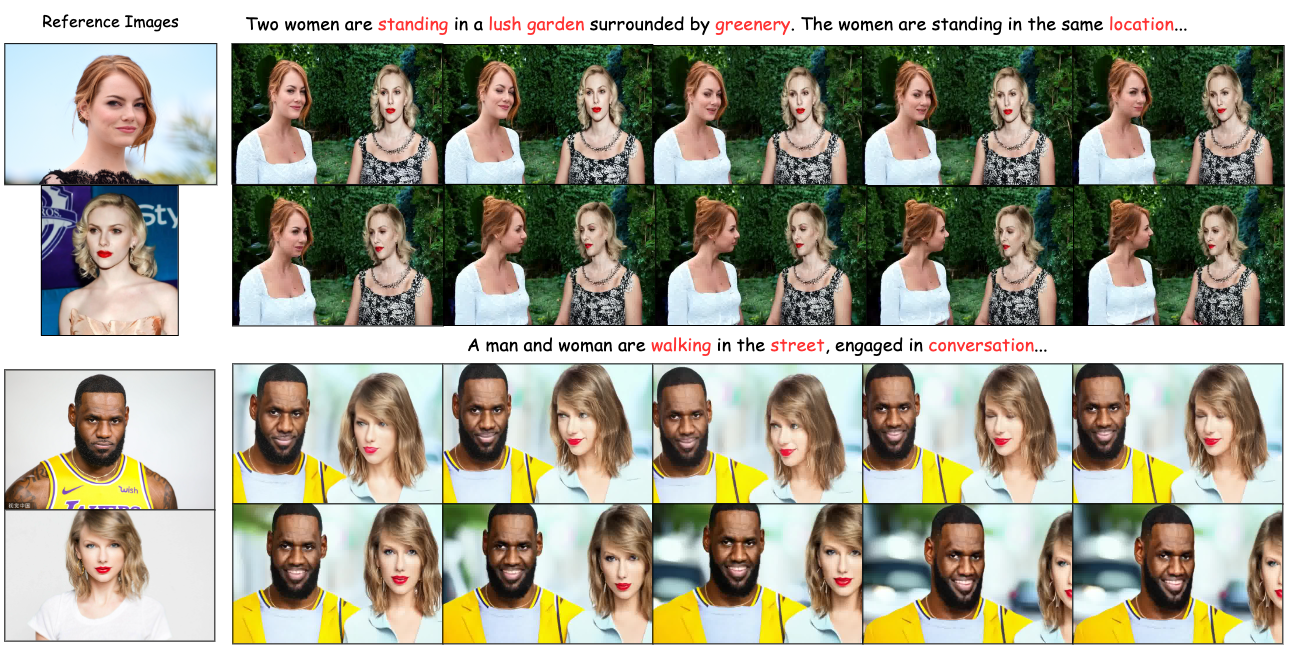}
   \caption{\textbf{Additional bad examples of multi-human customization.} Our \texttt{Ingredients} involves failures that generated characters appearing as though they were directly copied-pasted and out-painting, leading to an inconsistent video scenes.
   }
   \label{fig:add_example} 
\end{figure*}

\paragraph{ID Router.}

Through the projector, we derive a sequence of accurate facial embeddings for each ID. 
These embeddings are inherently position-independent. To prevent the blending of identities within the latent features, a position-wise routing network is employed to assign a unique identity to each potential facial region.
Let there be $N$ distinct human IDs in given image sets, with each ID facial embedding denoted as $ \tilde{F}^{n}$. 
For every tempera-spatial position $(x, y, t)$ at frame $t$ within the visual token $H \in \mathbb{R}^{c \times h \times w \times t}$ in the latent space, we route and assign a unique identity $k^* (0 \leq k^* \leq N-1)$ to it with $k^*$ determined as: 
\begin{equation}
\begin{split}
    k^* = \text{Argmax}_k \text{Softmax}(\phi(H, \tilde{F})), \\
    \phi(H, \Tilde{F}) = [f(H) * g(W_{aggr} *  \tilde{F}^{n})]_{n=1}^N,
\end{split}
\end{equation}
where $\phi$ represent router network that output routing logits as a $N$-dimensional discrete probability as $\mathbb{R}^{N \times h \times w \times t}$. 
$W_{aggr}$ is a learnable weight that aggregate multiple learnable latents from Q-former into a single token, $f(\cdot)$ and $g(\cdot)$ are two-layer MLP networks, $*$ is matrix multiplication operator. The idea behind the routing mechanism is that each patch $(x, y)$ frame $t$ in video is associated with at most one ID feature. 
We then gather the id feature set $\tilde{F} = \{\tilde{F}^1, \ldots, \tilde{F}^n\}$ according to index $k^*$ at each position to get the final mask ID feature $\overline{F}$, whose size is identical to queried visual tokens $H^{img}$. Finally, we obtain the refined visual tokens with residul connection as: 
\begin{equation}
    \Tilde{H} = H + \alpha_m * \overline{F},
\end{equation}
where $\alpha_m$ denote face scale to control facial feature influence.

\subsection{Training}
The training procedure is systematically divided into two distinct stages: facial embedding alignment phase and routing fine-tuning phase. During training, a random frame is selected from the training dataset that contains multi-ID videos, and Crop-and-Align operations are applied to isolate the facial regions corresponding to each identity, which then serve as reference control signals.

\paragraph{Facial Embedding Alignment.}

This phase focus on the face extractor and multi-scale projector. All inputs for the facial embedding are harnessed from the given reference image. Following previous study \cite{yuan2024identity,UniPortrait}, we also employ a dropping regularization, i.e., uniformly select global, local and completed facial information for video diffusion transformer condition. To enhance the integration of facial features, Low-Rank Adaptation (LoRA) \cite{hu2021lora} is appended to the video DiT architecture as additional trainable module. 
The training loss is aligned with conventional diffusion loss, as expressed in Equation \ref{eq: original_loss_function}.

\paragraph{Router Fine-Tuning.}
After completing facial embedding alignment, fine-tuning of ID router is conducted. We fix all the parameters except routing network as well as a LoRA of DiT. 
To supervise the correct routing, we assume video sequence contains distinct ID at each tempera-spacial patch. We first convert videos into masks $\mathcal{M}$, where -1 represents the background and $n$ denoted the corresponding ID region with face detection. 
The mask matrix $M$ is then down-sampled to match the size of video latents using \texttt{F.interpolate} with trilinear mode. The routing regularization can be optimized with a multi-label cross-entropy loss as:
\begin{equation}
    L_{route} = L_{diff} - \lambda \frac{1}{N} \sum_n M_n \text{log}(\phi(H, \tilde{F})_n),
\end{equation}
where $\lambda$ is the weight factor that balances routing loss with diffusion loss $L_{diff}$, $\phi((\phi(H, \tilde{F})$ is the routing logits $\mathbb{R}^{N \times h \times w \times t}$ capturing probabilities across all position in latent space. The mask $M \in \mathbb{R}^{1 \times h \times w \times t}$ serves as the ground truth for facial region labels, which is first used face detection to find boxes and then segmented with SAM \cite{kirillov2023segment}. Notably, indices of -1 for background class are excluded, ensuring the router focuses only on facial regions.

\section{Experiments}

\begin{table}[t]
\caption{\textbf{Compare between our method and baseline approaches on multiple human video generation.} Cogvideox served as the text-only baseline without any video conditioning while Inversion is tuning-based textual inversion. }
	\centering
    \begin{tabular}{c|ccc}		
Methods&  { Face  Sim. $\uparrow$ (\%) } & { CLIPScore $\uparrow$ (\%) } & {FID $\downarrow$}  \\
		\hline
	CogvideoX & 2.8 &28.3 & - \\ 
        Inversion & 35.6 & 24.3 & 154.2 \\
        Ingredients & 77.1 & 26.7 &106.3  \\ 
	\end{tabular}
	\label{tab:main}
\end{table}

\subsection{Experimental Setup}

\paragraph{Datasets.}
We utilize an in-house multi-human text-video dataset for training, which is curated from publicly available online resources and annotated with Qwen2-VL\footnote{https://huggingface.co/Qwen/Qwen2-VL-72B-Instruct} for detailed captioning. 
To mitigate the risk of domain bias, we ensure a balanced of topics throughout the dataset.
We use facexlib\footnote{https://github.com/xinntao/facexlib} to filter out and retain videos that contains human face number larger than two. 
As there is an absence of a multi-ID evaluation dataset, we select 50 images from diverse areas and racial categories, all excluded from the training corpus. We then design 60 distinct prompts, encompassing a variety of expressions, actions, and backgrounds. The image domains tree diagram and prompt that used to generate textual signals with GPT-4o are shown in Appendix. 
We also release all evaluation code as well as data in Huggingface for results reproduction and comparison.

\paragraph{Implementation Details.}

We select video diffusion transformers architecture CogvideoX-5B \cite{yang2024cogvideox} as baseline for performance validation. During training, video frames are processed at a resolution of 480$\times$720 , with 49 sequential frames extracted per video at an interval of two frames. 
The batch size is configured to 64, with a learning rate of 2e-5 at facial embedding alignment and 3e-5 at router fine-tuning. The total number training steps involves to 3k and 2k, respectively. AdamW is used as optimizer and cosine with restarts as scheduler. We set the face scale $\alpha_m$ to 1.0. During inference, we employ DPM \cite{DDPM} with a total of 50 sampling steps, and a text-guidance ratio of 6.0.

\paragraph{Evaluation Metrics.}

Following \cite{UniPortrait,yuan2024identity}, we assess the video customization across four dimensions: (\textbf{i}) Identity preservation: calculate pairwise face similarities between reference and generated faces with FaceNet \cite{schroff2015facenet}. Here, all detected faces in the generated frames are compared against reference faces through a greedy matching algorithm. The minimum similarity score across all matches is reported. (\textbf{ii}) Visual quality: we measure FID \cite{FID} by calculating feature differences in the face regions between generated frames and real face images within the InceptionV3 \cite{inceptionv3} features spaces. 
(\textbf{iii}) Text relevance: we assess with the average CLIP-B\footnote{https://huggingface.co/openai/clip-vit-base-patch32} image-text cosine similarity.

\subsection{Main Results}

We begin by assessing the general performance of multi-ID video generation with baselines. We examine the quality by using the 100 test cases from in-house test sets that randomly select two conformed texts for each images from evaluation sets. 
Table \ref{tab:main} provides a quantitative comparison of our proposed \texttt{Ingredients} approach against competitors. We can see that optimization-based techniques, that employing LoRA fine-tuning for each image pair \cite{textual_inversion}, often struggle to preserve individual identities, frequently producing generic or blended representations of the reference subjects. In contrast, the training-free \texttt{Ingredients} method demonstrates superior identity preservation, maintaining the distinct characteristics of each subject and achieving  higher automatic scores.
Moreover, our method exhibits prompt consistency comparable to text-only baselines, indicating that the personalization process does not compromise the original following capabilities of system. Qualitative results, presented in Figure \ref{fig:compare_tt}, further corroborate these findings. 

Additionally, Figure \ref{fig:add_example} highlights common failure cases observed in multi-ID video generation with varying prompts.
The first category of failure involves generated characters appearing as though they were directly copied and pasted, likely due to limitations in the image-to-video pattern initialization. The second category includes instances where the generated characters do not closely resemble the reference images, potentially caused by routing mis-classification. Addressing these issues remains an area for future research.

\begin{table}[t]
\caption{\textbf{Ablation studies for components in ID embedding and control signals.} Combine all of I2V initialization, segment supervision and spacal concatenation before VAE provide a best generative performance with multi-ID consistency.}
	\centering
    \small
    \begin{tabular}{cc|cc|cc|ccc}		
\multicolumn{2}{c|}{Initialize} &  \multicolumn{2}{|c|}{Router Supervised} & \multicolumn{2}{|c|}{VAE Features} & \multicolumn{3}{|c}{Metrics} \\ \hline
T2V & I2V &Box & Seg. &After & Before&
{ Face  Sim. $\uparrow$ (\%)} & { CLIP-T $\uparrow$ (\%)} & {FID $\downarrow$}  \\
		\hline
	$\checkmark$& & &$\checkmark$& &$\checkmark$ & 58.1 &26.5 & 122.5 \\ 
    &$\checkmark$ & &$\checkmark$&$\checkmark$ & &65.5 &25.9& 119.2 \\ 
    &$\checkmark$ &$\checkmark$ && &$\checkmark$ & 74.3 &26.7 &110.4 \\
    &$\checkmark$ & &$\checkmark$&& $\checkmark$ & 77.1 & 26.7 &106.3 \\
	\end{tabular}
    	\label{tab:abla}
\end{table}

\subsection{Ablation Study}

\paragraph{Effect of Model Initialize.}
There are usually two modes under general conditions, text-to-video (T2V) or image-to-video (I2V), for video generation models. Here, we evaluate both initialization methods while keeping the model architecture, learning parameters, and training steps consistent across experiments.
The results, presented in Table \ref{tab:abla}, reveal that initialization with image conditions yields notably higher facial similarity in the generated outputs compared to text-based initialization, as anticipated. However, the performance of both approaches is nearly identical with respect to adherence to textual prompts. Therefore, we use I2V as initialization be default.

\paragraph{Influence of Router Supervised Signals.}
To enable the ID router to accurately select relevant patch regions, we supervised its outputs with predefined signals. Specifically, we incorporate two types of signals: bounding boxes generated by a face detection model and segmented regions produced by the SAM model \cite{kirillov2023segment}. A detailed comparison of the hyperparameters used for face detection is provided in the Appendix.
The evaluated results, summarized in Table \ref{tab:abla}, indicate that signals derived from SAM yielded higher facial similarity scores. This improvement can be attributed to SAM's ability to better focus on facial regions while minimizing noise from the background, hair, and clothing.

\paragraph{How to Combine Global Embedding in VAE.}
In addition to integrating local features via routing, it is crucial to address the fusion of global features. We investigate two strategies for combining multiple reference images at the VAE level. Specifically, we examine whether the reference images should first be concatenated in pixel space and then processed by the VAE, or if the VAE should first process the images respectively and concatenate the features at a latent space. The results are presented in Table \ref{tab:abla}. Our findings demonstrate that concatenating the images before passing them through the VAE effectively enhances the naturalness of the generated results and improves the ability to following the textual input.

\paragraph{Effect of Routing Loss.}

\begin{table}[t]
	\caption{\textbf{Effect of routing loss.} Equipped with routing loss of ID classification helps to build a multi-ID consistent generation.}
	\label{tab:loss}
	\centering
    \begin{tabular}{c|ccc}		
Methods&  { Face  Sim. $\uparrow$ (\%) } & { CLIPScore $\uparrow$ (\%) } & {FID $\downarrow$}  \\
		\hline
	w/o $L_{route}$ & 62.2 & 26.9 & 112.3\\ 
        w/ $L_{mse}$ & 72.5 & 26.1 & 109.5 \\
        w/ $L_{route}$ & 77.1 & 26.7 & 106.3\\
	\end{tabular}
\end{table}

We finally test the routing loss design, including the omission routing loss and replace classification with MSE loss. 
Table \ref{tab:loss} provides a verification of the efficacy of the routing loss. The results suggest that classification loss can substantially enhance ID similarity while concurrently maintain prompt consistency in multi-human customization.

\subsection{Visualization}

\begin{figure*}[t]
  \centering
   \includegraphics[width=0.99\linewidth]{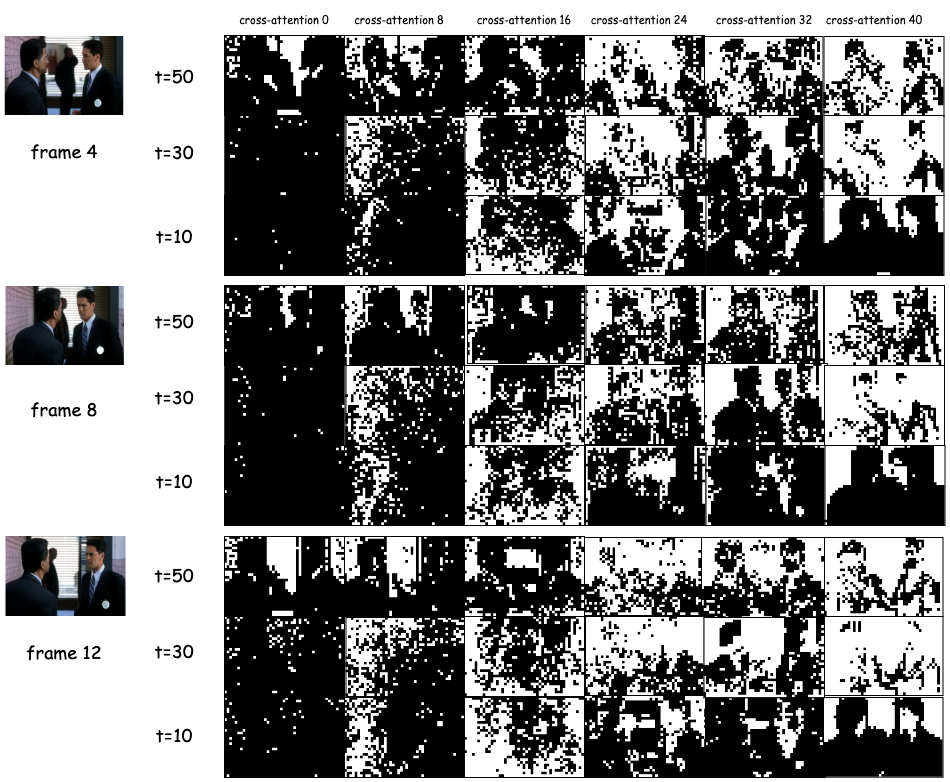}
   \caption{\textbf{Visualization of routing map within each cross-attention layer of video diffusion transformers.} We can see that with the routing loss, the routing network can discern different human IDs at earlier timesetps and in a more pronounced manner. 
   }
   \label{fig:routing} 
\end{figure*}

Figure \ref{fig:routing} illustrates the routing map for each cross-attention layer of \texttt{Ingredients} at different diffusion steps. 
In the visualization, white and black pixels represent two distinct human IDs. To enhance the clarity of the routing patterns, we upsample the argmax routing logits from the latent space. Every fourth frame is selected from the latent space, and every eighth cross-attention layer is displayed across varying denoising time steps.
The analysis reveals two key observations: (\textbf{i}) With the incorporation of routing loss, ID router can discern different human IDs at earlier timesteps, achieving increasingly refined distinctions at later denoising steps. That is, optimize the video generation in a fine-grained manner. 
(\textbf{ii}) At shallow cross-attention layers, the routing is chaotic and random. As the layers deepen, the routing pattern becomes more structured, eventually focusing on specific regions or semantic parts of the image. This progression aligns with the underlying principles of diffusion transformers, where the model gradually refines its visual representation over multiple steps.

\section{Conclusion}

We introduce \texttt{Ingredients}, a framework developed for multi-human customization based on video diffusion transformers. \texttt{Ingredients} incorporates a facial extractor that extract high-fidelity identity embedding for each ID, a multi-scale projector that combine facial embedding into context space of image query and an ID router to distribute and avoid identity blending. The empirical results demonstrate that \texttt{Ingredients} can deliver a synthesis that is not only of high quality and diversity but also offers robust editability and strong identity fidelity. We anticipate that our method will establish a new benchmark in the field, providing a replicable framework that can be adopted, extended, and optimized by future research efforts.

\bibliographystyle{ieee}
\bibliography{main}

\clearpage

\appendix

\begin{figure*}[t]
  \centering
   \includegraphics[width=0.8\linewidth]{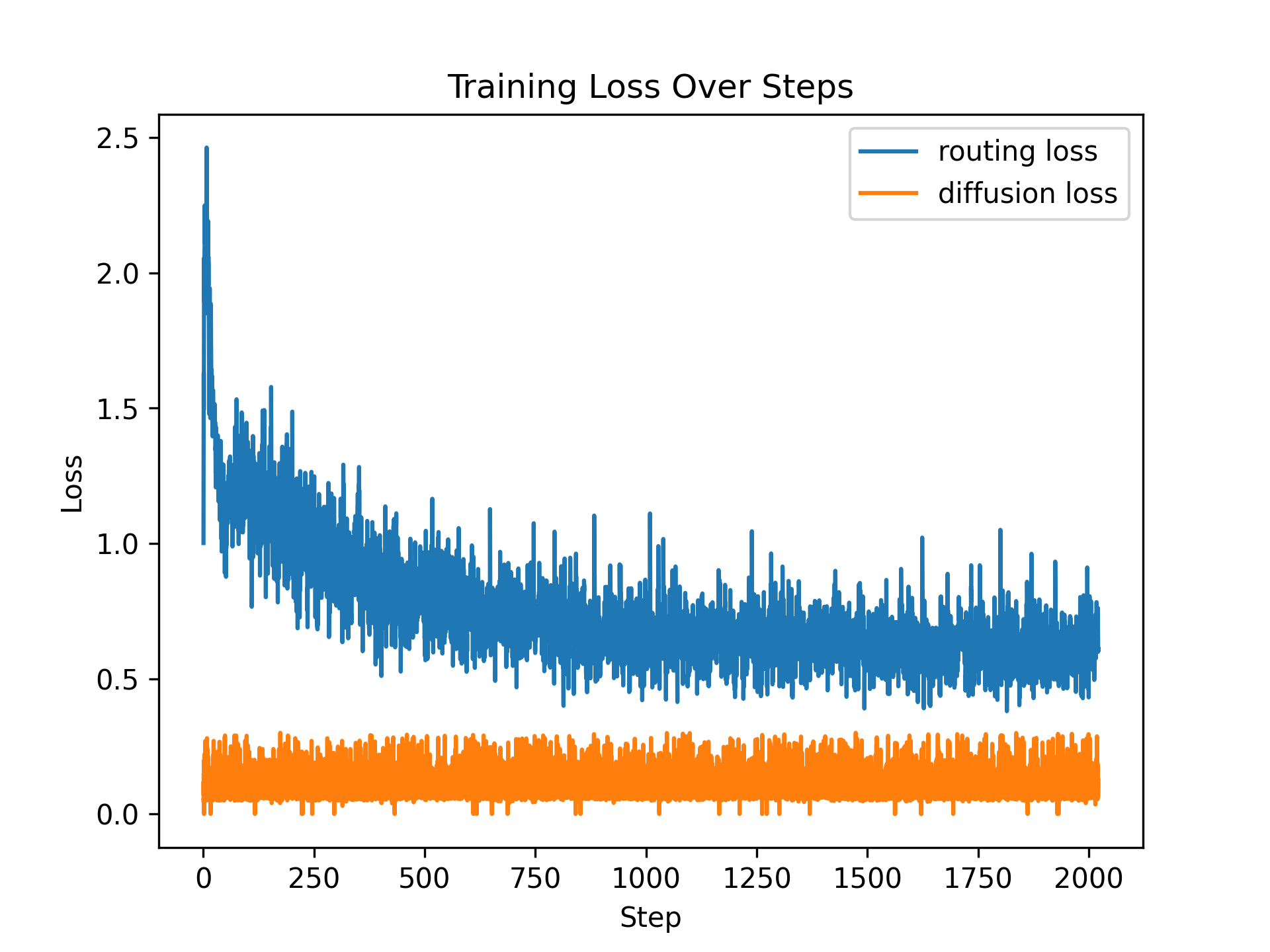}
   \caption{\textbf{The curve of different training loss in router fine-tuning stage.} We can see that with training steps increases, routing loss significantly decreases, the router becomes more accurate, while the diffusion loss remains almost unchanged, maintaining the original generative performance.
   }
   \label{fig:training_loss} 
\end{figure*}

\begin{figure*}[t]
  \centering
   \includegraphics[width=0.99\linewidth]{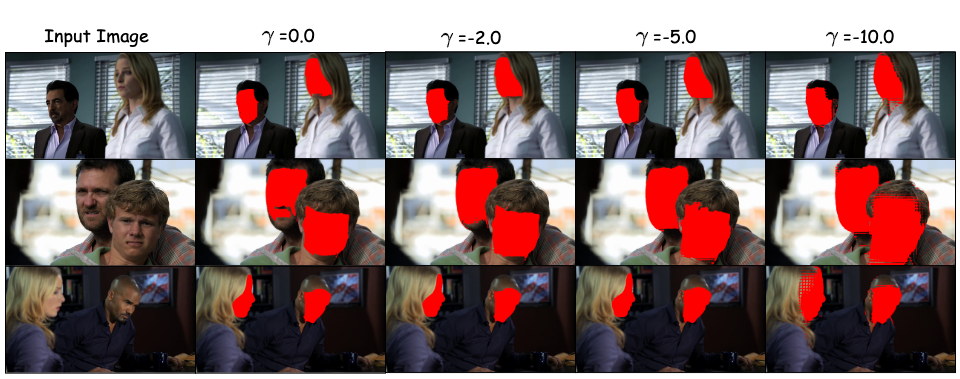}
   \caption{\textbf{Hyper-parameter settings for SAM segmentaion.} We select -2.0 as threshold to build routing supervised labels.
   }
   \label{fig:hyper} 
\end{figure*}

\begin{figure*}[t]
  \centering
   \includegraphics[width=0.99\linewidth]{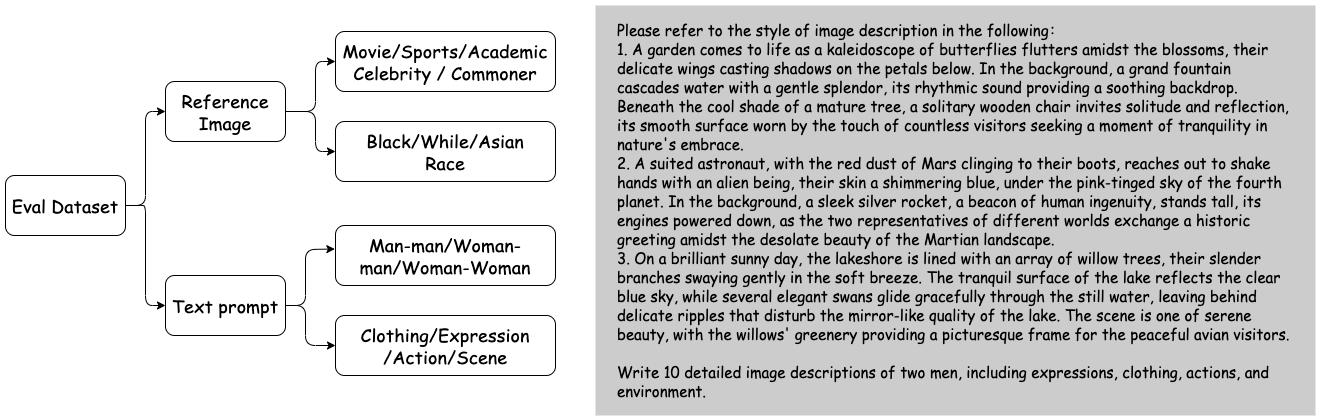}
   \caption{\textbf{Domain distribution of evaluation images (left) and used prompt to generate text inputs (right).} We consider multiple aspects for data collection to make evaluation more robust.
   }
   \label{fig:evaluate} 
\end{figure*}

\end{document}